\documentclass[acmtog,nonacm]{acmart}

\usepackage{enumitem}
\usepackage{subcaption}
\usepackage{makecell}

\copyrightyear{2026}
\setcopyright{cc}
\setcctype{by-nc-nd}
\begin{document}
\title{TailorRig: Controllable and Structured Auxiliary Skeleton Generation for Garment-Aware Rigging}
\author{Yajing Chen}
\authornote{Yajing Chen, Xin Chen, and Saidan Lu contributed equally to this research.}

\orcid{0000-0003-1146-2334}
\email{jadeyjchen@tencent.com}
\affiliation{%
  \institution{LIGHTSPEED, Tencent}
  \city{Shenzhen}
  \country{China}}

\author{Xin Chen}
\authornotemark[1]
\orcid{0009-0002-8152-3809}
\email{xinc3292@gmail.com}
\affiliation{%
  \institution{Xi'an Jiaotong University}
  \city{Xi'an}
  \country{China}}

\author{Saidan Lu}
\authornotemark[1]
\orcid{0009-0001-4498-8618}
\email{saidanlu@tencent.com}
\affiliation{%
  \institution{LIGHTSPEED, Tencent}
  \city{Shanghai}
  \country{China}}

\author{Yeqi Yuan}
\orcid{0009-0003-9445-5523}
\email{2401210510@pku.edu.cn}
\affiliation{%
  \institution{Peking University}
  \city{Beijing}
  \country{China}}

\author{Weikai Chen}
\orcid{0000-0002-3212-1072}
\email{chenwk891@gmail.com}
\affiliation{%
  \institution{LIGHTSPEED, Tencent}
  \city{Los Angeles}
  \country{United States}}

\author{Changyu Hu}
\orcid{0009-0003-7173-8328}
\email{hucy24@mails.tsinghua.edu.cn}
\affiliation{%
  \institution{Tsinghua University}
  \city{Beijing}
  \country{China}}

\author{Cheng Zeng}
\orcid{0009-0006-9121-454X}
\email{zeng-c24@mails.tsinghua.edu.cn}
\affiliation{%
  \institution{Tsinghua University}
  \city{Beijing}
  \country{China}}

\author{Xin Wang}
\orcid{0000-0002-6789-4569}
\email{alexinwang@tencent.com}
\affiliation{%
  \institution{LIGHTSPEED, Tencent}
  \city{Shenzhen}
  \country{China}}

\renewcommand{\shortauthors}{Chen et al.}

\begin{teaserfigure}
  \includegraphics[width=\textwidth]{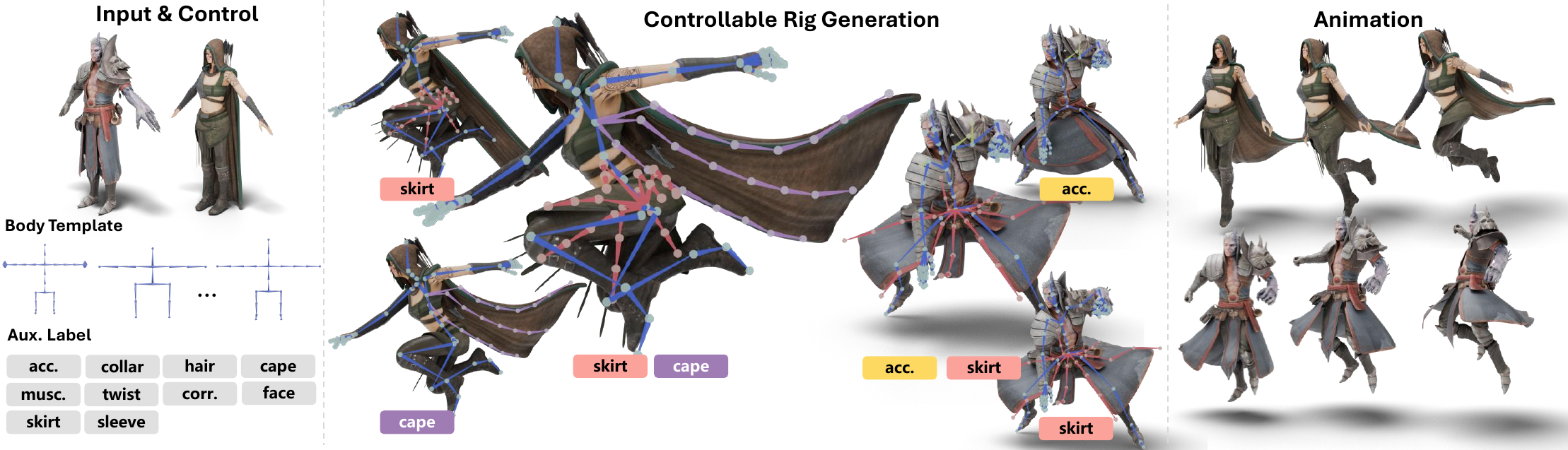}
  \caption{\textbf{TailorRig turns artist controls into animation-ready garment-aware rigs.}
  TailorRig accepts any body template and any subset of auxiliary category labels (\emph{left}). Different label choices yield distinct full rigs (\emph{middle}, with chosen labels marked under each result), all driving stable animation (\emph{right}).
  3D characters courtesy of Adobe Mixamo~\cite{adobe2021mixamo}.}
  \Description{The teaser image presents a three-stage pipeline for character rigging.}
  \label{fig:teaser}
\end{teaserfigure}

\begin{abstract}
Humanoid rigging for modern characters with complex garments remains challenging due to the need for dense auxiliary skeletons and precise skinning control. 
Existing automatic methods primarily recover rigs from geometry, providing limited artist control over auxiliary articulation and often lacking explicit semantic or kinematic-chain structure for dense garment bones, leading to instability and weight bleeding.
We present TailorRig, a framework that reformulates garment-aware humanoid rigging as a structured, controllable full-rig generation problem.
This formulation shifts the goal from automatic rig recovery to artist-driven rig authoring, where both body-template compatibility and auxiliary articulation are explicit design choices.
TailorRig exposes two explicit axes of artist control within a unified framework: a user-specified body skeleton template that preserves body-rig compatibility, and artist-provided semantic labels that drive an autoregressive generator to synthesize requested category-specific auxiliary bone chains.
To robustly assign skinning weights under dense skeletal configurations, we further introduce a hierarchical chain-to-bone formulation: vertices are first associated to a kinematic chain, and bone weights are then refined within that chain, reducing weight bleeding between spatially adjacent but semantically unrelated structures.
We evaluate TailorRig on a diverse set of characters and introduce Human-Mix, a large-scale dataset of over 38,000 humanoid models curated for garment-aware rigging. Experiments show that TailorRig improves skeleton accuracy, deformation stability, controllability, and practical usability over existing methods.
Project page: \url{https://xxxx0601.github.io/trig/}. Code: \url{https://github.com/Tencent/TailorRig}.
Dataset: \url{https://huggingface.co/datasets/tencent/TailorRig}.
\end{abstract}

\maketitle

\section{Introduction}
\label{sec:intro}

Humanoid rigging is essential for articulated deformation in animation pipelines, including games, films and virtual production
~\cite{parent2012computer}.
The recent surge in generative 3D assets~\cite{hong2023lrm,li2023instant3d,liu2023one,poole2022dreamfusion,siddiqui2024meshgpt,zhang2024clay,lai2025hunyuan3d} requires scalable and automated rigging solutions.
While significant progress has been made in automating body rigging, modern characters increasingly feature complex garments, accessories, and layered structures, such as skirts, capes, coats, and long hair, that require auxiliary skeletal structures to deform naturally. 
These auxiliary bones are no longer optional embellishments, but essential components of production-oriented character rigs.

A large body of prior work has sought to automate humanoid rigging. 
Template-based approaches~\cite{baran2007pinocchio,guo2025mia,chu2025humanrig,ma2023tarig} leverage predefined skeletons to ensure compatibility with existing animation libraries, while template-free methods~\cite{zhang2025unirig,song2025magicarticulate,song2025puppeteer,lin2021point2skeleton,xu2020rignet} 
infer skeletal structures directly from geometry to improve flexibility and generalization.
However, when applied to characters with complex garments and dense 
auxiliary joints, these methods exhibit two recurring failures.
Auxiliary bones are treated \emph{implicitly}---either omitted altogether or 
generated indiscriminately---leaving artists little control over their 
presence, category, or placement.
Moreover, dense auxiliary skeletons exacerbate skinning ambiguity, 
causing weight bleeding across unrelated body and garment parts (e.g., a skirt erroneously following arm motion).
These failures share a common root: prior methods formulate rigging as \emph{geometry-to-rig recovery}, while production demands \emph{artist-driven rig authoring}.
We therefore reformulate garment-aware humanoid rigging as a \textbf{\emph{structured, controllable full-rig generation}} problem.
By \emph{controllable}, we mean that artists can specify both the body rig template and the auxiliary bone categories; by \emph{structured}, we mean that rig generation and skinning are guided by semantic grounding and kinematic-chain hierarchy.

Given an input mesh, an arbitrary user-defined body skeleton template, and optional semantic labels specifying auxiliary categories, TailorRig produces an animation-ready rig that preserves body skeleton compatibility while supporting editable auxiliary structures.
TailorRig consists of two core components (Fig.~\ref{fig:pipeline}).
\textbf{(1) Controllable body-to-auxiliary skeleton generation.}
A semantically-guided alignment module first maps an arbitrary user-defined body skeleton to the input character while preserving its kinematic topology, enabling immediate reuse of existing animation assets.
Conditioned on artist-provided semantic labels (e.g.\ skirt), a label-driven autoregressive generator then synthesizes category-specific auxiliary bone chains only when desired, providing the explicit control required in professional workflows.
\textbf{(2) Hierarchical chain-to-bone skinning.}
When dozens of garment bones from different chains overlap in close spatial proximity, a vertex cannot be correctly skinned by reasoning about individual bones in isolation.
We therefore first predict coarse vertex-to-chain associations -- identifying which kinematic branch (e.g., left-arm chain vs.\ skirt chain) each vertex belongs to -- and then refine per-bone weights within the assigned branch.
This two-level decomposition suppresses weight bleeding across unrelated body and garment parts.
Both components use a hierarchical bone encoder with stage-specific parameters, providing consistent kinematic priors across the pipeline.

To support training and evaluation, we curate Human-Mix, a large-scale dataset comprising over $38,000$ humanoid models with semantic annotations for garment-aware rigging.
We will release Human-Mix in a license-compliant form, sharing the in-house VRoid assets and annotations while providing annotations and source identifiers for retained third-party characters (Sec.~\ref{sec:data_release}).

In summary, our main contributions are:
\begin{itemize}[leftmargin=15pt, topsep=2pt, itemsep=1pt, parsep=0pt]
\item A controllable rigging formulation that supports explicit body template control and auxiliary label control, enabling artists to govern both body and garment skeletal structures.
\item A structured full-rig generation framework that injects semantic grounding and kinematic-chain hierarchy into body alignment, auxiliary-bone generation, and skinning.
\item A hierarchical chain-to-bone skinning method that reduces weight bleeding through hierarchical structural reasoning, where vertices are first localized to the correct kinematic branch before competing among nearby bones.
\item Human-Mix, a large-scale dataset of 38K+ humanoid models with semantic annotations for garment-aware rigging.
\end{itemize}
\noindent We validate these claims through dedicated controllability experiments, structural ablations targeting both semantic and geometric priors, weight bleeding analyses, and production-oriented evaluations on complex real-world assets.

\section{Related Work}

Existing methods generally fall into two categories: template-based and template-free approaches.

\subsection{Template-based Rigging.}
Template-based methods leverage a predefined skeletal topology to provide strong structural priors.
Pinocchio~\cite{baran2007pinocchio} pioneered this line by formulating joint placement as discrete geometric embedding on a medial surface, while later optimization approaches extended it to inverse-kinematics fitting~\cite{jacobson2011bounded,james2005skinning}; both families, however, typically require watertight meshes and offer limited adaptability.
Deep-learning methods then took over:
TARig~\cite{ma2023tarig} introduces cross-attention maps that regress vertex-to-joint correspondences for arbitrary mesh resolution, and MIA~\cite{guo2025mia} proposes a coarse-to-fine estimation strategy that progressively refines joint positions through multi-stage mesh-aware prediction.
Other concurrent works explore blendshape-coupled alignment~\cite{li2021neuralblendshapes}, image-lifted keypoints~\cite{chu2025humanrig}, and voxel-based prediction~\cite{lefevre2025comparing}.
Despite their robustness, these models are typically confined to a single skeletal standard, and their support for auxiliary bones~\cite{ma2023tarig,guo2025mia} remains limited by fixed topologies, failing to accommodate diverse clothing styles.

\subsection{Template-free Rigging.}
Template-free methods infer arbitrary topologies directly from geometry.
Classical approaches extract skeletons via medial-axis analysis~\cite{amenta1998surface,tagliasacchi2009curve}, while RigNet~\cite{xu2020rignet} pioneered learning-based rigging by predicting joints and connectivity with a graph neural network, followed by MoRig~\cite{xu2022morig} and HeterSkinNet~\cite{pan2021heterskinnet} for motion-aware and heterogeneous settings; Anymate~\cite{deng2025anymate} scales this predictive paradigm with a large rigged-asset corpus and a three-stage joint--connectivity--skinning framework.
Inspired by autoregressive 3D generation~\cite{ibing2023octree,yan2022shapeformer,mittal2022autosdf,nash2020polygen}, recent methods reformulate skeleton synthesis as sequence modeling: RigAnything~\cite{liu2025riganything} applies diffusion-guided BFS prediction, UniRig~\cite{zhang2025unirig} autoregressively decodes hierarchical skeletons from discretized spatial tokens, and MagicArticulate~\cite{song2025magicarticulate}, Puppeteer~\cite{song2025puppeteer}, and Auto-Connect~\cite{guo2025autoconnect} explore related grid-token formulations.
Though flexible, these generative approaches introduce topological stochasticity that often violates studio standards, and their implicit auxiliary generation~\cite{zhang2025unirig} lacks explicit semantic conditioning for artist-driven control.
More fundamentally, existing methods provide limited explicit modeling of garment semantics and kinematic-chain structure, which becomes critical when dense auxiliary bones must be generated and skinned under artist control, or for intersection-free garment retargeting~\cite{huang2025intersection}.

\subsection{Automatic Skinning.}
Traditional skinning methods rely on geometric formulations such as bounded-biharmonic weights~\cite{jacobson2011bounded}, geodesic voxel binding~\cite{dionne2013geodesic}, and heat-diffusion solvers \cite{baran2007pinocchio}, which assume watertight meshes and degrade when bone influences densely overlap on garment-rich characters; example-based methods instead approximate LBS from animated mesh sequences~\cite{james2005skinning,kavan2010fast}, which requires pre-existing animations.
Learning-based approaches improve quality substantially.
NeuroSkinning~\cite{liu2019neuroskinning} trains a global PointNet-based model for fixed-bone weight inference, while other template-specific methods~\cite{ma2023tarig,chu2025humanrig,guo2025mia} achieve high fidelity within their skeletal standards.
For arbitrary skeletons, graph-based architectures~\cite{xu2020rignet,pan2021heterskinnet} propagate bone influences along mesh edges, and transformer-based methods~\cite{song2025magicarticulate,guo2025autoconnect,zhang2025unirig,song2025puppeteer} model long-range vertex--bone interactions via attention; orthogonally, weight-transfer methods inpaint an existing weight set onto a new mesh~\cite{abdrashitov2023robust}.
Despite these advances, weight bleeding remains challenging for dense auxiliary bones, where closely packed bones create ambiguous influence boundaries.
Our hierarchical chain-to-bone skinning addresses this by first localizing each vertex to the correct kinematic chain, then refining per-bone weights within that chain, effectively suppressing cross-chain bleeding.

\section{Methods}

\begin{figure*}[t]
    \centering
    \includegraphics[width=\textwidth]{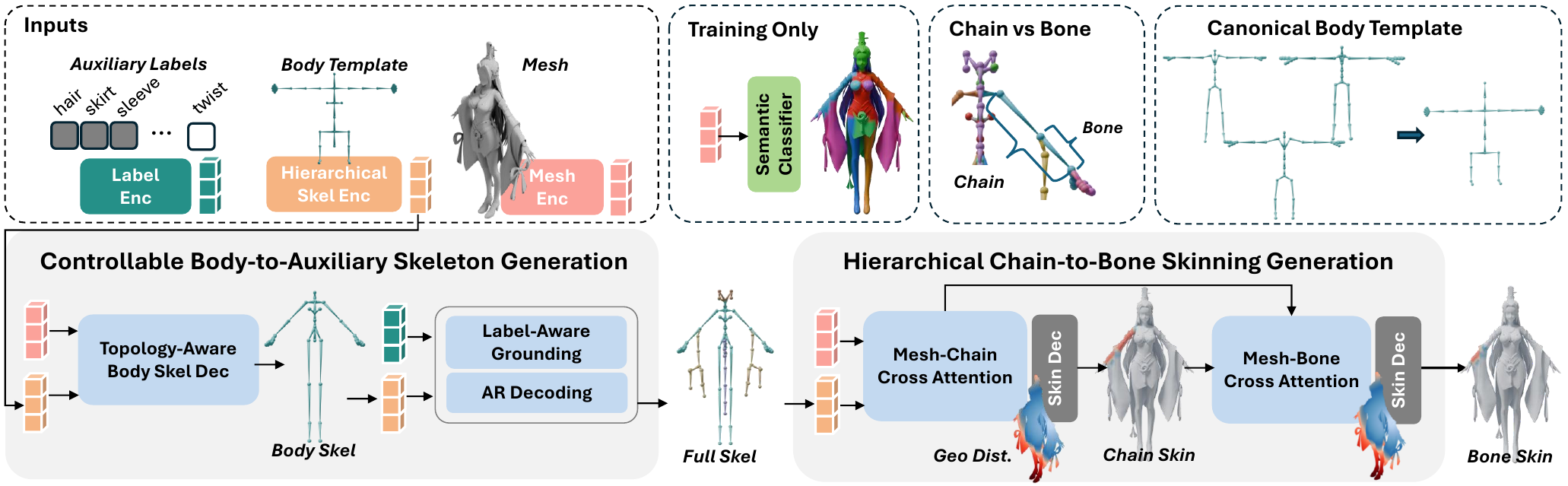}
    \caption{\textbf{Overview of TailorRig.}
    Given a mesh, a body skeleton template, and auxiliary category labels, TailorRig produces an animation-ready rig in two stages. 
    (1) \textbf{Controllable skeleton generation}: a Topology-Aware body decoder (TAMB) aligns the template to the mesh, and a Label-Aware autoregressive decoder synthesizes category-specific auxiliary chains, both use the same hierarchical skeleton-encoder architecture with stage-specific parameters.
    (2) \textbf{Hierarchical chain-to-bone skinning}: vertices are first assigned to kinematic chains and then refined into per-bone weights, suppressing cross-chain weight bleeding.}
    \Description{Pipeline diagram of TailorRig showing two stages: controllable skeleton generation that aligns a body template and synthesizes auxiliary bone chains, followed by hierarchical chain-to-bone skinning weight prediction.}
\label{fig:pipeline}
\end{figure*}

\subsection{Problem Formulation}
\label{sec:formulation}

We formulate garment-aware humanoid rigging as a
\textbf{controllable, structured full-rig generation} problem.
Given a mesh $\mathcal{M}=(\mathcal{V},\mathcal{E}_{\mathcal{M}})$
with vertex positions $\mathbf{V}^p$ and normals $\mathbf{V}^n$,
a user-specified body skeleton template
$\tilde{\mathcal{B}}_{\text{body}}$, and a set of auxiliary category
labels, TailorRig outputs a full skeleton
$\mathcal{B}=(\mathbf{J},\mathcal{E}_{\text{joint}})$ (body plus
garment-aware auxiliary bones) and skinning weights $\mathbf{W}$.
The two control inputs $(\tilde{\mathcal{B}}_{\text{body}},\text{labels})$
realize the \emph{controllable} axis; the \emph{structured} axis is
carried by a shared 21-class semantic taxonomy and a hierarchical
skeleton encoder (stage-specific parameters) that propagates
joint--chain structure into every prediction stage
(Sec.~\ref{sec:encoders}).
Built on these ingredients, the pipeline addresses three production
needs: template alignment for animation reuse
(Sec.~\ref{sec:alignment}), label-conditioned generation for
explicit auxiliary control (Sec.~\ref{sec:aux_skeleton}), and
chain-to-bone skinning against cross-branch weight bleeding
(Sec.~\ref{sec:skinning}).

\subsection{Hierarchical Skeleton Encoder and Common Modules}
\label{sec:encoders}

The following modules are used across skeleton and skinning
prediction. Each is defined by a common architectural design but
instantiated with stage-specific parameters: mesh, label, and bone
encoders, plus a semantic classifier that supervises a shared
per-vertex taxonomy during training (see Fig.~\ref{fig:pipeline}).

\paragraph{Mesh and label encoders.}
We employ a PartField-style encoder~\cite{liu2025partfield} to build a continuous 3D feature field from the input mesh.
Global mesh tokens $\mathbf{F}_{\mathcal{M}} \in \mathbb{R}^{N_{\mathcal{M}} \times D}$ are obtained by querying this field at $N_{\mathcal{M}}{=}512$ farthest-point-sampled locations, providing a compact shape summary used as mesh conditioning for skeleton generation;
per-vertex features $\mathbf{F}_{\mathcal{V}} \in \mathbb{R}^{V \times D}$ are obtained by evaluating the field at every mesh vertex, providing dense local geometry for bone encoding, semantic classification, and skinning prediction.
For each label, we map its category index to a learnable embedding vector $\mathbf{f}_{\text{label}} \in \mathbb{R}^D$, which is later used to query mesh- and body-aware features for label-conditioned auxiliary skeleton generation.

\paragraph{Hierarchical skeleton encoder.}
This module produces the per-joint representation
$\mathbf{F}_{\mathcal{B}} \in \mathbb{R}^{J \times D}$ by encoding the
skeleton at both joint and chain granularities.
Each joint coordinate is embedded with Fourier features and refined with
mesh context from $\mathbf{F}_{\mathcal{M}}$ to obtain mesh-aware joint tokens.
Each maximal unbranched path in the skeleton tree defines a kinematic chain
(e.g., spine, left arm, skirt-left), whose constituent tokens are aggregated
into a chain-level representation.
Fusing the joint- and chain-level features yields $\mathbf{F}_{\mathcal{B}}$,
so each joint reflects both local geometry and chain-level structure.
The same encoder architecture is instantiated at each stage---on the template
body skeleton for alignment, on the aligned body skeleton for auxiliary generation,
and on the complete skeleton for skinning---providing a consistent two-level
structural prior throughout the pipeline.

\paragraph{Semantic classifier.}
A vertex-wise classifier predicts a distribution over 21 semantic classes
(11 body kinematic-chain classes and 10 auxiliary classes) from
$\mathbf{F}_{\mathcal{V}}$.
It is instantiated at each stage with stage-specific parameters and label masks
(11 body classes for alignment; all 21 classes for auxiliary generation and skinning),
providing consistent semantic grounding across body alignment, auxiliary generation,
and skinning.

\subsection{Topology-Aware Body Skeleton Generation}
\label{sec:alignment}

The body skeleton template serves as TailorRig's first explicit control signal. 
By accepting arbitrary artist-specified templates rather than fixed or stochastic skeletons, this module aligns each template to the target mesh while preserving kinematic topology and semantic consistency, enabling immediate reuse of existing animations.

\paragraph{Canonical Template Space}

For an arbitrary input template, we first convert it to a standard T-pose and then normalize it in a unified geometric space to reduce differences in pose, orientation, and global scale (see Fig.~\ref{fig:pipeline}). After canonicalization, the template preserves its topology while being mapped into a consistent coordinate system shared with the target mesh.

\paragraph{Topology-Aware Body Skeleton Decoder}

To generate body joint positions that are both geometrically accurate and topologically consistent, we propose a Topology-Aware Mesh-Bone (TAMB) fusion decoder. We apply $L$ layers of topology-aware mesh--bone fusion blocks with pre-norm residual connections. At each layer $l$, we first perform cross-attention from bones to mesh and then refine via EdgeConv~\cite{wang2019edgeconv} over the bone connectivity graph:
\begin{align}
\mathbf{F}^{(l)}_{\text{attn}} &= \mathbf{F}^{(l-1)}_{\mathcal{B}} + \mathrm{Attn}\!\big(\mathrm{LN}(\mathbf{F}^{(l-1)}_{\mathcal{B}}),\, \mathbf{F}_{\mathcal{M}},\, \mathbf{F}_{\mathcal{M}}\big),\label{eq:tamb_attn}\\
\mathbf{F}^{(l)}_{\mathcal{B}} &= \mathbf{F}^{(l)}_{\text{attn}} + \mathrm{MLP}\!\big(\mathrm{EdgeConv}(\mathrm{LN}(\mathbf{F}^{(l)}_{\text{attn}}),\, \mathcal{E}_{\text{body}})\big).\label{eq:tamb_edge}
\end{align}
Here, $\mathrm{Attn}(\mathrm{Q}, \mathrm{K}, \mathrm{V})$ denotes multi-head cross attention and $\mathrm{LN}(\cdot)$ denotes layer normalization. The decoder is initialized with the template body skeleton features, i.e., $\mathbf{F}^{(0)}_{\mathcal{B}} = \mathbf{F}_{\tilde{\mathcal{B}}_{\text{body}}}$. After $L$ layers, the fused bone features are decoded into aligned joint positions via an MLP head.

\paragraph{Semantic and Geometric Supervision}
\label{sec:align_loss}

The overall body alignment loss combines position, projection, and semantic terms:
$
\mathcal{L}_{\text{align}}=
\mathcal{L}_{\text{pos}}
+\mathcal{L}_{\text{proj}}
+\mathcal{L}_{\text{sem}}.
$
$\mathcal{L}_{\text{pos}}$ is a standard $L_1$ regression loss on joint positions.
$\mathcal{L}_{\text{sem}}$ is a vertex-wise classification loss (11 body kinematic-chain classes) that enforces mesh--bone semantic correspondence, ensuring each predicted joint is grounded to its correct body region.
The key geometric term is the \textbf{bone projection loss} $\mathcal{L}_{\text{proj}}$, designed to suppress anatomically implausible transverse drift.
For joint $i$ with parent $p(i)$, we project $\hat{\mathbf{j}}_i$ onto the ground-truth bone segment $[\mathbf{j}_{p(i)}, \mathbf{j}_i]$ and penalize the squared distance to the closest point:
$
\mathcal{L}_{\text{proj}} = \frac{1}{J} \sum_{i=1}^{J} \big\| \hat{\mathbf{j}}_i - \mathbf{p}_{\text{close}, i} \big\|_2^2,
$
where $\mathbf{p}_{\text{close}, i}$ is the closest point on the GT bone segment.
This lets the predicted joint slide along the bone direction while penalizing lateral deviation, preventing drift off its kinematic axis.

\subsection{Label-Aware Auxiliary Skeleton Generation}
\label{sec:aux_skeleton}

Auxiliary bone labels are the second explicit control signal in TailorRig. Given a category label (e.g., ``skirt''), the model generates auxiliary bones for that category, anchored to the mesh region. We model this as an autoregressive generation problem conditioned on mesh and body features and a category label.

\paragraph{Label-to-Geometry Semantic Grounding}

Rather than using the category label as a static global condition, we ground it in the current mesh and body-skeleton context through a multi-step attention mechanism.
We first apply one round of reciprocal cross-attention with pre-norm residual connections between global mesh tokens $\mathbf{F}_{\mathcal{M}}$ and body-skeleton features $\mathbf{F}_{\mathcal{B}_{\text{body}}}$ to obtain mutually aligned features $\mathbf{F}_{\mathcal{M}}'$ and $\mathbf{F}_{\mathcal{B}_{\text{body}}}'$.
We then use the label embedding $\mathbf{f}_{\text{label}}$ as a query to probe both aligned feature sets via cross-attention, producing two geometry-aware label tokens---one encoding where the category manifests on the mesh surface, the other on the body skeleton.
These tokens are concatenated with the aligned features to form the conditioning prefix $\mathbf{F}_{\text{prefix}} \in \mathbb{R}^{(N_{\mathcal{M}}+J_{\text{body}}+2) \times D}$ for the autoregressive decoder.

\paragraph{Structured Autoregressive Decoding}

Following~\cite{zhang2025unirig}, we serialize each auxiliary skeleton tree into a token sequence via depth-first traversal, with joint coordinates quantized into discrete bins. We adopt OPT-350M~\cite{zhang2022opt} as the decoder backbone. The conditioning prefix $\mathbf{F}_{\text{prefix}}$ is treated as token embeddings and prepended to the skeleton token sequence, allowing the decoder to attend to mesh and body context through its standard causal self-attention. The model then autoregressively predicts the auxiliary bone sequence.

\paragraph{Loss Functions}

The auxiliary generation module is trained with a next-token prediction loss over the skeleton token sequence, supplemented by the same semantic classification loss $\mathcal{L}_{\text{sem}}$ (all 21 classes) to enhance semantic consistency and label-aware grounding.

\subsection{Hierarchical Chain-to-Bone Skinning Generation}
\label{sec:skinning}

We implement the skinning stage as a chain-to-bone decomposition:
vertices are first softly assigned to kinematic chains, and
per-bone weights are then refined under this chain assignment,
restricting bone-level competition within each chain.

\paragraph{Chain-level representation.}
We define a kinematic chain as a maximal unbranched path of consecutive bones between two branching or terminal joints in the skeleton tree. The full skeleton (body + auxiliary) is automatically partitioned into $K$ such chains (e.g., spine, left arm, skirt-left), where $K$ varies per character depending on the skeleton topology. Each chain representation is obtained by mean-pooling constituent bone features from the hierarchical skeleton encoder, yielding chain features $\mathbf{F}_{\mathcal{C}} \in \mathbb{R}^{K \times D}$.

\paragraph{Hierarchical feature refinement.}
Feature interaction proceeds in a coarse-to-fine manner via reciprocal cross-attention. In the coarse stage, vertex features and chain features alternately serve as query and key/value for 4 rounds of cross-attention, yielding chain-aware vertex features $\mathbf{F}_{\mathcal{V}}^{c}$ and vertex-aware chain features $\mathbf{F}_{\mathcal{C}}'$. In the fine stage, vertex features and bone features undergo the same reciprocal interaction for 8 rounds, producing refined bone-aware vertex features $\mathbf{F}_{\mathcal{V}}^{b}$ and vertex-aware bone features $\mathbf{F}_{\mathcal{B}}'$. The coarse stage requires fewer rounds as chain-level associations are simpler than bone-level ones.

\paragraph{Chain-to-bone weight prediction.}
The chain weight decoder predicts coarse chain weights $\mathbf{W}_{\mathcal{C}} \in \mathbb{R}^{V \times K}$ by combining two cues for each vertex--chain pair: a learned similarity from cross-attention between vertex and chain features, and a geometric prior derived from the geodesic distance between the vertex and the nearest bone in each chain.
The bone weight decoder follows a similar design but additionally conditions on the predicted chain weight: each bone's parent-chain weight is concatenated into its feature as a structural prior, which is combined with the bone-level similarity and distance cues to predict the final per-bone weights $\mathbf{W} \in \mathbb{R}^{V \times B}$.
Because the parent-chain weight acts as a continuous structural prior, bone-level competition is effectively concentrated within each vertex's associated chains, suppressing cross-chain weight bleeding.

\paragraph{Loss Functions}
\label{sec:skin_loss}

The overall skinning objective combines four terms:
$
\mathcal{L}_{\text{skin\_total}}
= \lambda_{\text{skin}} \mathcal{L}_{\text{skin}}
+ \lambda_{\text{def}} \mathcal{L}_{\text{def}}
+ \lambda_{\text{chain}} \mathcal{L}_{\text{chain}}
+ \lambda_{\text{sem}} \mathcal{L}_{\text{sem}},
$
where $\mathcal{L}_{\text{skin}}$ is the KL divergence between ground-truth and predicted per-bone weight distributions, $\mathcal{L}_{\text{def}}$ measures vertex-level deformation error under $N{=}3$ random training LBS poses, $\mathcal{L}_{\text{chain}}$ supervises the coarse chain-level weight distribution (chain weights are obtained by summing per-bone weights within each chain), and $\mathcal{L}_{\text{sem}}$ is the same vertex-wise semantic classification loss used in body alignment (Sec.~\ref{sec:align_loss}), here applied with all 21 classes to suppress cross-part bleeding.

\section{Experiments}
\label{sec:experiments}

We evaluate TailorRig on rigging quality (Sec.~\ref{sec:main_comp}), controllability (Sec.~\ref{sec:ctrl_eval}), structural ablations (Sec.~\ref{sec:struct_val}), and in-the-wild production readiness (Sec.~\ref{sec:in_the_wild}).

\subsection{Experimental Setup}
\label{sec:exp_setup}

\begin{figure}[!ht] 
    \centering 
    \includegraphics[width=0.95\linewidth]{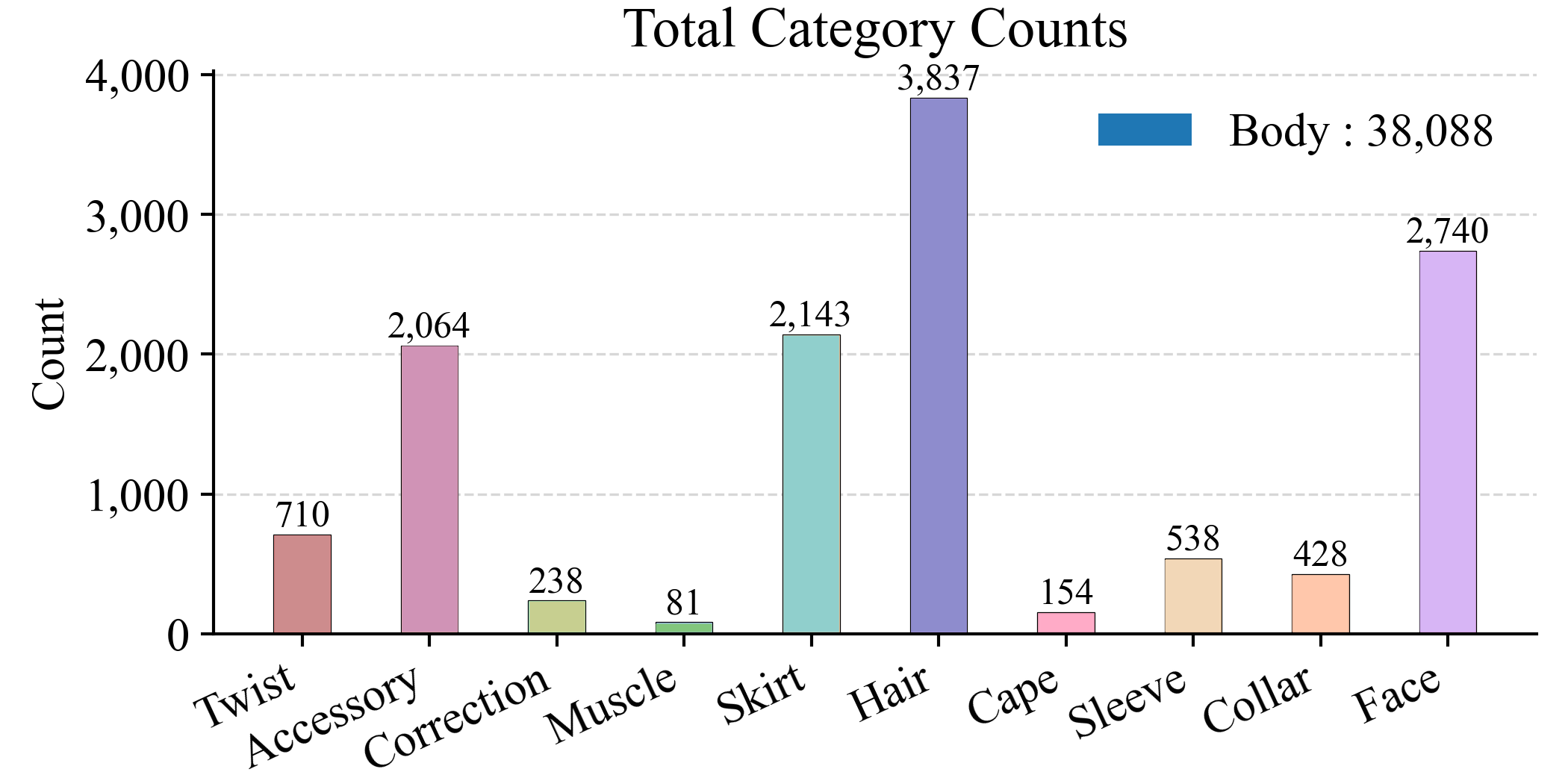} 
    \caption{Auxiliary-class distribution in Human-Mix. Bars report per-class counts across the 10 auxiliary classes among the 7,824 characters containing at least one auxiliary bone; the \emph{Body} group (11 body kinematic-chain classes) covers all 38,088 characters and is reported in the legend.} 
    \Description{Bar chart showing the per-class distribution of the ten auxiliary bone categories in the Human-Mix dataset, alongside eleven body kinematic-chain classes that cover all characters.}
\label{fig:dataset_hist} 
\end{figure}

\paragraph{Dataset.}
No existing dataset provides auxiliary bone annotations for training and evaluating TailorRig.
We therefore construct \textbf{Human-Mix}, aggregating 38,088 high-quality rigged humanoid characters from three public sources and an in-house VRoid collection (Table~\ref{tab:dataset_detailed}) with diverse skeletal structures and semantic annotations, spanning a broad spectrum of auxiliary bone configurations.
All characters undergo rigorous multi-stage curation and are annotated with our 21-class bone taxonomy (11 body, 10 auxiliary) co-designed with professional artists, of which 7,824 contain at least one auxiliary class (Fig.~\ref{fig:dataset_hist}).
All Mixamo-derived assets are excluded from training via source-independent geometric de-duplication (Sec.~\ref{sec:data_release}); they appear only in qualitative demonstrations.

\begin{table*}[t]
\centering
\caption{Source composition and curation statistics of \textbf{Human-Mix}. \#Raw: characters collected before curation; \#Curated: characters retained after curation; \#Aux.: characters containing at least one auxiliary bone category.}
\Description{Table listing the source composition and curation statistics of the Human-Mix dataset.}
\label{tab:dataset_detailed}
\small
\begin{tabular}{lccc}
\toprule
\textbf{Source} & \textbf{\#Raw} & \textbf{\#Curated} & \textbf{\#Aux.}  \\
\midrule
Articulation-XL-v2~\cite{song2025magicarticulate} & 48,635 & 22,533 & 4,060  \\
HumanRig~\cite{chu2025humanrig}\textsuperscript{$\dagger$}           & 11,434 & 11,434  & 0        \\
VRoid\textsuperscript{$*$}~\cite{isozaki2021vroid}                      & 3,404  & 3,404  & 3,337  \\
ModelResource~\cite{xu2020rignet}      & 2,703  & 717    & 427  \\
\midrule
\textbf{Total (Human-Mix)} & --\textsuperscript{$\ddagger$} & \textbf{38,088} & \textbf{7,824} \\
\bottomrule
\end{tabular}

{\raggedright\footnotesize
\textsuperscript{$*$}Created in-house using VRoid Studio; used under its content license.\\
\textsuperscript{$\dagger$}HumanRig is a pre-curated dataset; we adopt it in full without additional filtering.\\
\textsuperscript{$\ddagger$}Raw totals are omitted because cross-source overlap prevents simple summation.\\
We will release the in-house VRoid subset as rigged FBX assets with annotations; for retained third-party characters we release our annotations with original source identifiers (Sec.~\ref{sec:data_release}).\par}

\end{table*}

\paragraph{Data provenance and release.}
\label{sec:data_release}
We shortlist cross-source duplicate candidates by vertex and bone counts, then manually confirm matches with mean post-ICP vertex error below 0.001 using skeleton topology and joint distances as additional checks. The same source-independent procedure is applied against a Mixamo reference library; matched assets are excluded before training. The 3,404 VRoid characters were created in-house using VRoid Studio under its content license. Upon publication, we plan to release these rigged assets with our annotations. For retained characters from Articulation-XL-v2, HumanRig, and ModelResource, we plan to release only our annotations and original source identifiers, allowing users to obtain the assets from their original providers. Mixamo assets are used for qualitative figures and are not included in the training data or dataset release.

\paragraph{Baselines.}
For their applicable tasks, we retrain UniRig, MIA, RigNet, and NBS on the same Human-Mix split using released pipelines, changing only the data loader. Methods without reproducible training pipelines use official checkpoints, while non-learning baselines require no retraining.

\paragraph{Metrics.}
\textbf{Skeleton geometry.} We measure skeletal fidelity via bidirectional 
Chamfer Distance at three granularities: CD-J2J (joint-to-joint), 
CD-J2B (joint-to-bone segment), and CD-B2B (bone-to-bone stick). 
For auxiliary bones, a predicted bone inherits the category of a GT bone when it 
lies inside that bone's capsule---a cylinder of radius 0.15 around 
the joint-to-joint segment---and the three distances are computed 
within each category.
\textbf{Skinning.} We report the $L_1$ error of the skinning matrix 
and Deformation $L_1$ ($\mathcal{E}_{\text{def}}$), which evaluates 
vertex-level geometric distortion under LBS with $N{=}5$ random test poses.

\paragraph{Implementation details.}
Each mesh is resampled to $16{,}384$ points and normalized to a unit ball; we use a $90\%{:}5\%{:}5\%$ split with random augmentation.
The feature dimension is $D{=}512$; the skeleton and skinning modules are trained independently with AdamW ($\text{LR}{=}10^{-5}$, weight decay $0.05$).

\begin{table*}[t]
\centering
\small
\renewcommand{\arraystretch}{0.95}
\setlength{\tabcolsep}{4pt}
\caption{Quantitative comparison across all rigging tasks in TailorRig. Values in Skinning Weight are scaled by $10^3$.}
\Description{Table of quantitative comparisons with baseline methods across all rigging tasks, reporting skeleton and skinning metrics.}
\label{tab:combined_comp_results}

\begin{tabular}{l ccc ccc cc}
\toprule
& \multicolumn{3}{c}{\textbf{Body Skeleton Alignment}} & \multicolumn{3}{c}{\textbf{Auxiliary Bone Generation}} & \multicolumn{2}{c}{\textbf{Skinning Weight}} \\
\cmidrule(lr){2-4} \cmidrule(lr){5-7} \cmidrule(lr){8-9}
\textbf{Method} & J2J$\downarrow$ & J2B$\downarrow$ & B2B$\downarrow$ & J2J$\downarrow$ & J2B$\downarrow$ & B2B$\downarrow$ & $L_1 \downarrow$ & Def-$L_1 \downarrow$ \\
\midrule
MIA~\cite{guo2025mia}         & 0.065 & 0.053 & 0.031 & --   & --    & --   & 14.43 & 1.56 \\
TARig~\cite{ma2023tarig}      & 0.061 & 0.045 & 0.034 & --    & --    & --    & 28.25 & 2.68 \\
NBS~\cite{li2021neuralblendshapes} & 0.141 & 0.121 & 0.111 & -- & -- & -- & 62.86 & 28.52 \\
RigNet~\cite{xu2020rignet}    & 0.147 & 0.111 & 0.100 & 0.141 & 0.127 & 0.118 & 8.56  & 2.14 \\
MagicArti.~\cite{song2025magicarticulate} & 0.064 & 0.046 & 0.031 & 0.202 & 0.209 & 0.197 & --\textsuperscript{$\|$}    & --\textsuperscript{$\|$}   \\
Puppeteer~\cite{song2025puppeteer} & 0.062 & 0.044 & 0.032 & 0.201 & 0.211 & 0.198 & 10.31 & 0.64 \\
RigAnything~\cite{liu2025riganything} & 0.060 & 0.046 & 0.030 & 0.170 & 0.174 & 0.168 & --\textsuperscript{$\S$}    & --\textsuperscript{$\S$}   \\
UniRig~\cite{zhang2025unirig} & 0.040 & 0.031 & 0.020 & 0.191 & 0.209 & 0.219 & 11.55 & 0.91 \\
GVB~\cite{dionne2013geodesic} & --    & --    & --    & --    & --    & --    & 26.67 & 2.08 \\
Pinocchio~\cite{baran2007pinocchio} & 0.101    & 0.072    & 0.045    & --    & --    & --    & 48.77 & 4.42 \\
\midrule
\textbf{Ours}                 & \textbf{0.023} & \textbf{0.013} & \textbf{0.011} & \textbf{0.068} & \textbf{0.096} & \textbf{0.098} & \textbf{7.70} & \textbf{0.60} \\
\bottomrule
\end{tabular}

{\raggedright\footnotesize
\textsuperscript{$\S$}RigAnything jointly predicts skeleton and skinning and does not accept an external skeleton; controlled skinning is therefore inapplicable.\\
\textsuperscript{$\|$}MagicArticulate shares the same skinning module as Puppeteer; results are not duplicated.\par}
\end{table*}

\begin{table*}[!thp]
\centering
\small
\renewcommand{\arraystretch}{0.95}
\setlength{\tabcolsep}{5pt}
\caption{Ablation study on body skeleton alignment, auxiliary bone generation, and skinning quality. Values in Skinning Quality are scaled by $10^3$.}
\Description{Table of ablation results on body skeleton alignment, auxiliary bone generation, and skinning quality.}
\label{tab:ablation_all}

\begin{tabular}{l ccc ccc cc}
\toprule
& \multicolumn{3}{c}{\textbf{Body Skeleton Alignment}} & \multicolumn{3}{c}{\textbf{Auxiliary Bone Generation}} & \multicolumn{2}{c}{\textbf{Skinning Weight}} \\
\cmidrule(lr){2-4} \cmidrule(lr){5-7} \cmidrule(lr){8-9}
\textbf{Variant} & J2J$\downarrow$ & J2B$\downarrow$ & B2B$\downarrow$ & J2J$\downarrow$ & J2B$\downarrow$ & B2B$\downarrow$ & $L_1 \downarrow$ & Def-$L_1 \downarrow$ \\
\midrule

Ours w/o $\mathcal{L}_{\text{sem}}$ & 0.029 & 0.018 & 0.015 & 0.120 & 0.130 & 0.131 & 11.50 & 0.83 \\
Ours w/o Hier.\ Encoder & 0.026 & 0.015 & 0.013 & 0.096 & 0.106 & 0.102 & 18.60 & 1.88 \\
Ours w/o TAMB Fusion & 0.030 & 0.018 & 0.015 & -- & -- & -- & -- & -- \\
Ours w/o $\mathcal{L}_{\text{proj}}$ & 0.031 & 0.020 & 0.017 & -- & -- & -- & -- & -- \\
Ours w/o Label-Aware Attn & -- & -- & -- & 0.107 & 0.126 & 0.125 & -- & -- \\
Ours w/o Chain-to-Bone & -- & -- & -- & -- & -- & -- & 11.24 & 0.84 \\
\midrule
\textbf{Ours} & \textbf{0.023} & \textbf{0.013} & \textbf{0.011} & \textbf{0.068} & \textbf{0.096} & \textbf{0.098} & \textbf{7.70} & \textbf{0.60} \\
\bottomrule
\end{tabular}

{\raggedright\footnotesize
Each variant removes one component from the full model. ``--'' marks task groups unaffected by the ablated component: e.g., TAMB Fusion and $\mathcal{L}_{\text{proj}}$ act only on body alignment; Label-Aware Attn acts only on auxiliary generation; Chain-to-Bone acts only on skinning.\par}
\end{table*}

\subsection{Comparison Studies}
\label{sec:main_comp}

\begin{figure*}[p]
    \centering
    \includegraphics[width=1.0\linewidth]{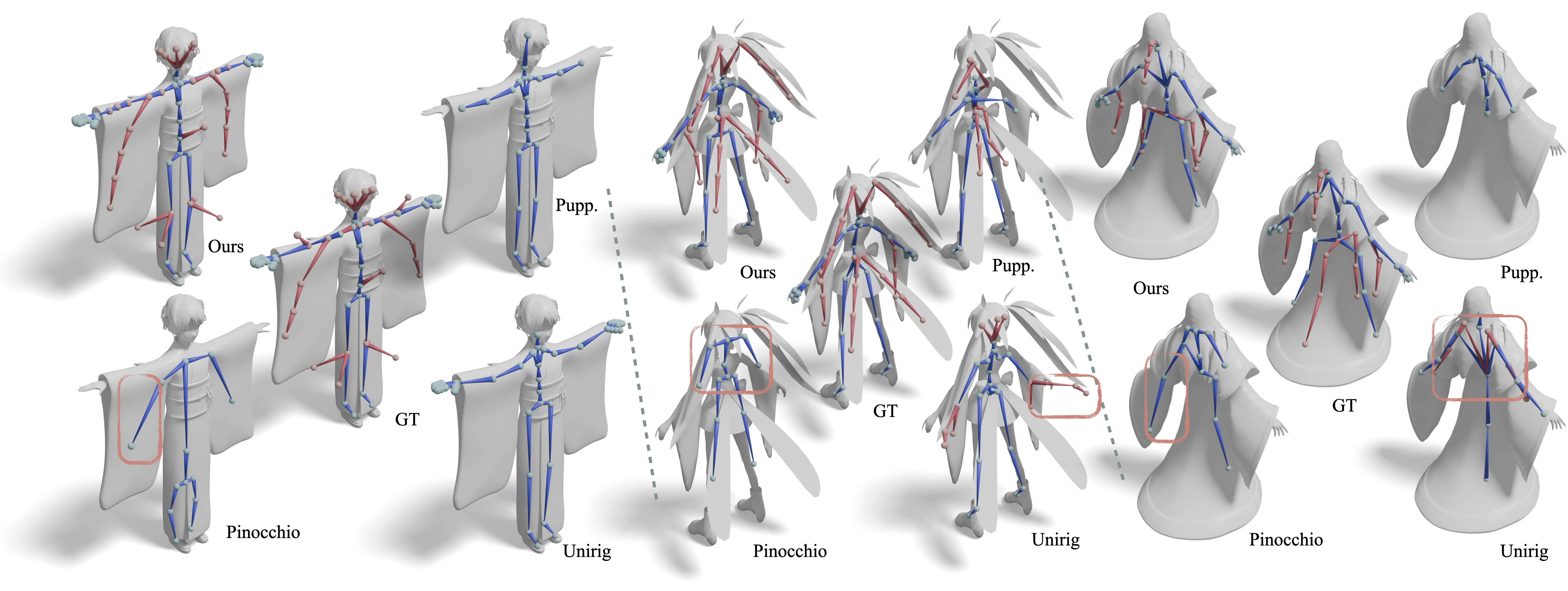}
    \caption{\textbf{Skeleton comparison on garment-rich characters.}
    Blue denotes body bones, and red denotes auxiliary or garment-related bones
    (manually colored for baselines when no explicit auxiliary labels are provided).
    Baselines often omit, merge, or mis-parent garment-related auxiliary bones,
    whereas TailorRig produces cleaner body--auxiliary separation under the specified labels
    and better matches the ground truth.}
\Description{Grid of garment-rich characters comparing skeletons from TailorRig and baseline rigging methods, with body bones in blue and auxiliary bones in red, highlighting cleaner body-auxiliary separation from TailorRig.}
\label{fig:skeleton_comparison}

\end{figure*}

\begin{figure*}[p]
    \centering
    \includegraphics[width=1.0\linewidth]{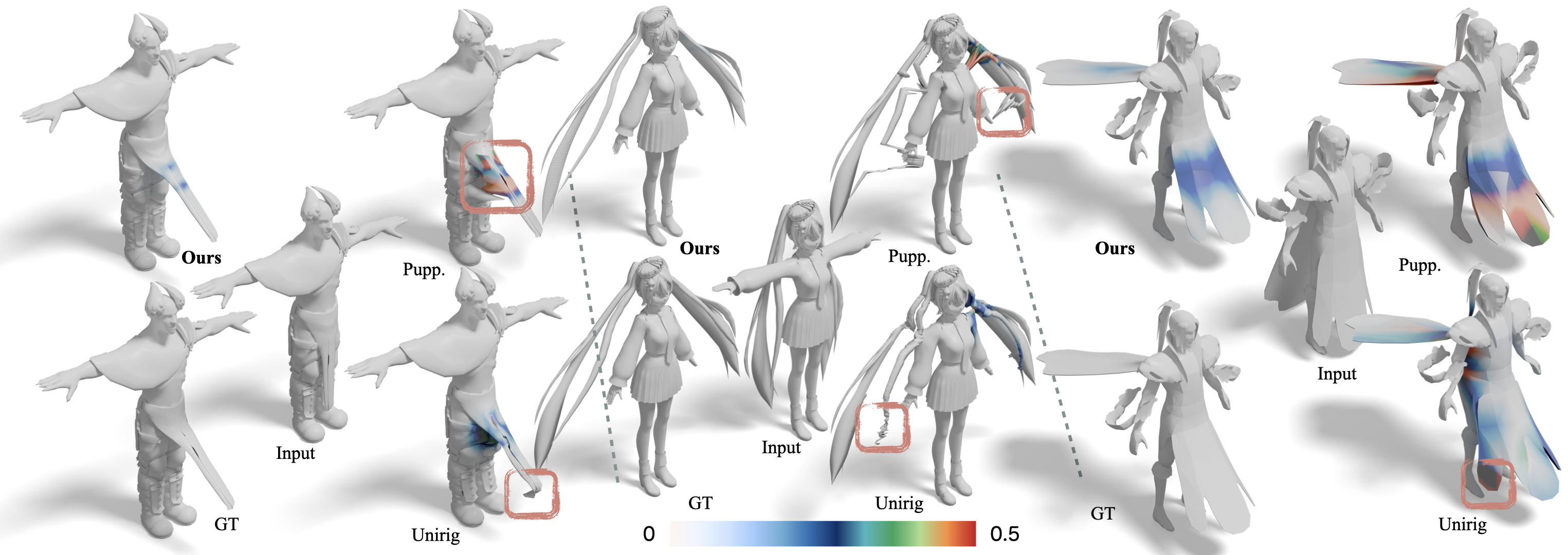}
    \caption{\textbf{Skinning comparison under controlled auxiliary-bone rotations.}
    All methods use the same ground-truth skeleton and are evaluated under the same auxiliary-bone rotations, isolating skinning-weight prediction from skeleton errors.
    Surface heat maps visualize skinning-weight error against the ground truth, with warmer colors indicating larger errors.
    TailorRig produces cleaner garment/accessory deformation, whereas baselines exhibit stronger body--auxiliary coupling and local artifacts.}
    \Description{Deformation comparisons under identical auxiliary-bone rotations, with surface heat maps of skinning-weight error; TailorRig shows mostly cool colors (low error) while baselines show warmer error regions.}
\label{fig:skinning_comparison}
\end{figure*}

\begin{figure*}[p]
    \centering
    \includegraphics[width=1.0\linewidth]{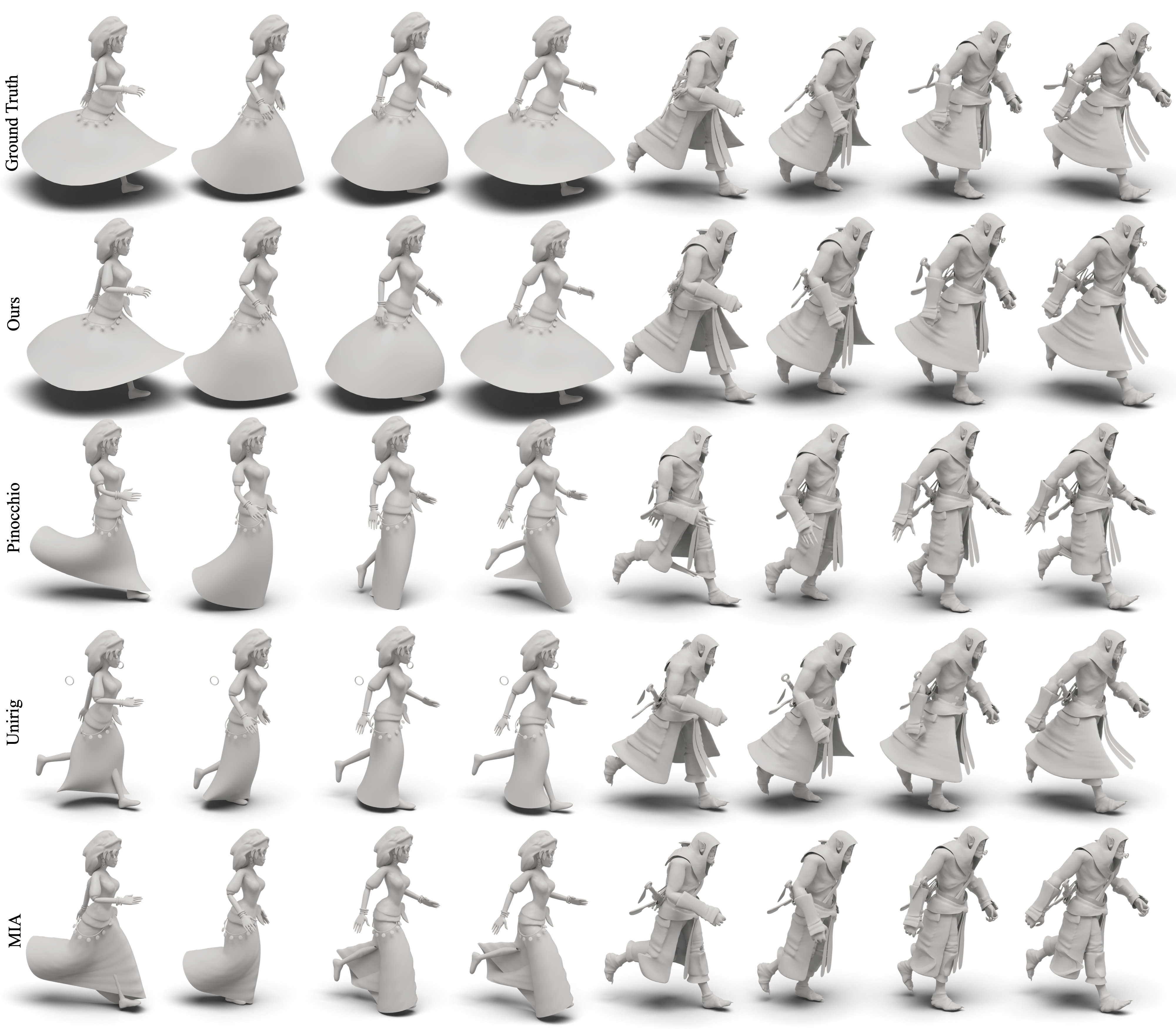}
    \caption{\textbf{Frame-wise animation comparison after full rigging.}
    After each method generates a complete skeleton and skinning weights, we drive all results with the same motion and render the same sampled frames.
    TailorRig better preserves garment motion relative to the body, closely matching the ground truth.
    In contrast, Pinocchio and UniRig exhibit garment--body coupling, and UniRig further shows leg--cloth interpenetration in the highlighted regions.}
    \Description{Sampled animation frames driven by the same motion across rigging methods; TailorRig preserves garment motion close to the ground truth, while baselines show garment-body coupling and leg-cloth interpenetration.}
\label{fig:animation_comparison}
\end{figure*}

Table~\ref{tab:combined_comp_results} reports quantitative results and Figs.~\ref{fig:skeleton_comparison}--\ref{fig:animation_comparison} provide qualitative comparisons.
Our method achieves the best score on every metric.

\paragraph{Skeleton quality.}
For body skeletons, TailorRig achieves J2J\,0.023, improving on the best
template-based baseline (TARig, 0.061) and reducing UniRig's 0.040 by 42.5\%;
UniRig is template-free and thus serves as a reference rather than a
same-input baseline.
For auxiliary bones, the gain is larger: J2J decreases from UniRig's 0.191 to 0.068, indicating that label-conditioned
generation better localizes garment-aware chains.
Fig.~\ref{fig:skeleton_comparison} further illustrates that existing methods often
confuse garment-related auxiliary bones with the body skeleton.
For example, Pinocchio inserts body bones into the sleeve region in the left
example and the hair region in the middle example, while learning-based
baselines produce missing or incorrectly attached auxiliary chains.
TailorRig avoids these failures by using the specified labels, yielding cleaner
body--auxiliary separation and topology closer to the ground truth.

\paragraph{Skinning quality.}
For skinning, we evaluate all applicable methods in a controlled setting using the same ground-truth skeleton, thereby isolating skinning-weight prediction from skeleton errors.
TailorRig achieves the best skinning scores in Table~\ref{tab:combined_comp_results}, with scaled $L_1$ / Def-$L_1$ errors of 7.70 / 0.60.
As shown in Fig.~\ref{fig:skinning_comparison}, this advantage is most visible under auxiliary-bone rotations: baselines exhibit stronger body--auxiliary coupling and local artifacts, such as braid--body adhesion and skirt coupling in the highlighted regions, whereas TailorRig keeps the deformation closer to the ground truth.

\paragraph{Full-rig animation.}
We further evaluate the complete rigs under the same input motion.
As shown in Fig.~\ref{fig:animation_comparison}, prior methods exhibit garment--body
coupling and interpenetration artifacts, while TailorRig's dedicated auxiliary
structures and cleaner skinning produce more stable frame-wise deformation.
The sampled frames in Fig.~\ref{fig:animation_comparison} illustrate the resulting garment motion.

\begin{figure}[!b]
    \centering
    \includegraphics[width=1\linewidth]{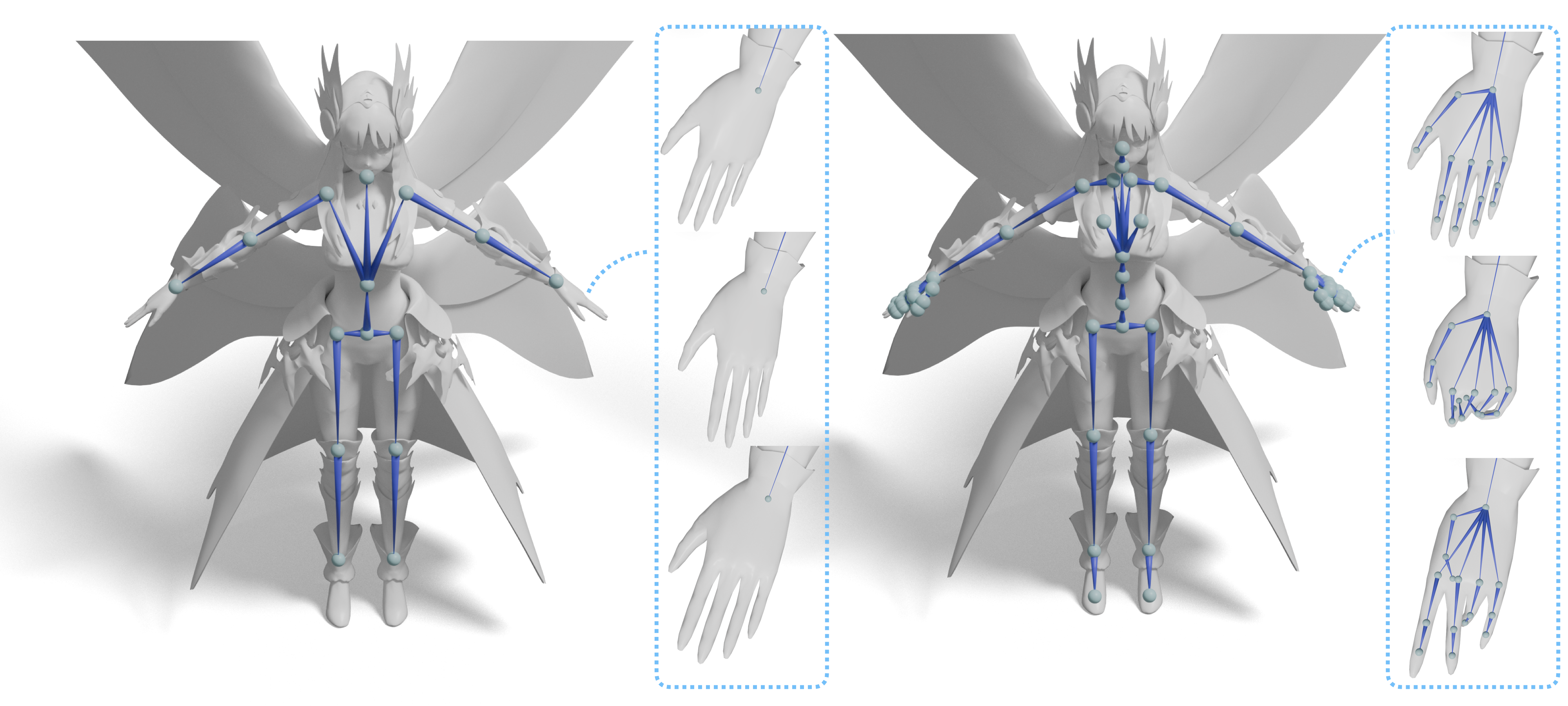}
    \caption{\textbf{Body-template control.} Different body templates control different rigs: a template without finger joints (left) yields a finger-less hand rig, while one with articulated fingers (right) yields an articulated one—producing distinct hand behaviors under the same motion.}
    \Description{Side-by-side comparison of hand rigs generated from two body templates: a template without finger joints produces a finger-less hand rig, while one with articulated fingers produces an articulated hand, shown under the same driving motion.}
\label{fig:controllability_body_finger}
\end{figure}

\begin{figure*}[t]
    \centering
    \includegraphics[width=\linewidth]{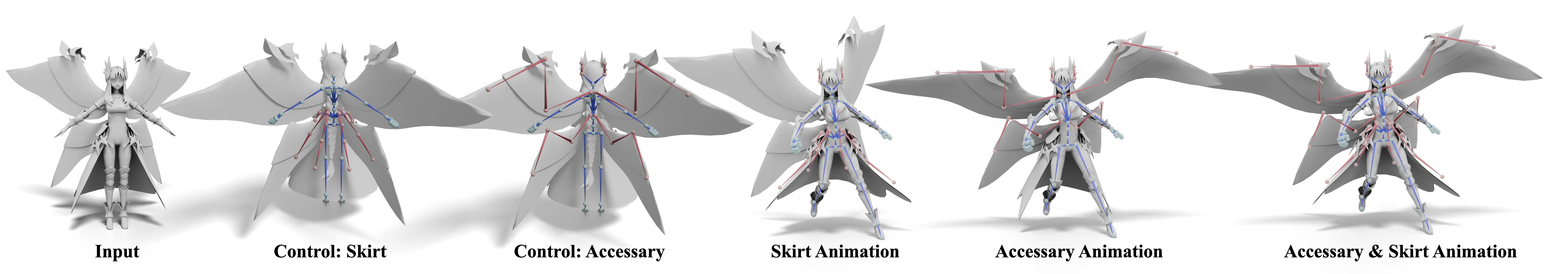}
    \caption{\textbf{Auxiliary-label control.} With body skeleton and mesh fixed, different auxiliary labels control the generated bones and, under the same driving motion, lead to different animations across the \emph{skirt}, \emph{accessory}, and \emph{skirt+accessory} rigs. \textcolor{blue}{Blue}/\textcolor{red}{red}: body/auxiliary bones.}
    \Description{Rendered comparisons showing how different auxiliary label selections (skirt, accessory, or both) change the generated bones and the resulting garment animation under the same motion, with body bones in blue and auxiliary bones in red.}
\label{fig:controllability}
\end{figure*}

\subsection{Controllability Evaluation}
\label{sec:ctrl_eval}

\paragraph{Body-template control.}
Fig.~\ref{fig:controllability_body_finger} shows that different body templates produce distinct rigs and hand behaviors on the same mesh, confirming template-level structural control.
\paragraph{Auxiliary-label control.}
Fig.~\ref{fig:controllability} shows that different labels activate only the corresponding auxiliary bones and produce distinct animated behaviors.
Quantitatively, we request one GT auxiliary category per test character and report two metrics:
requested-bone recall, the fraction of requested-category GT bones matched by same-category predictions under the capsule rule of Sec.~\ref{sec:exp_setup};
and off-label rate, the fraction of predicted auxiliary bones whose category falls outside the request.
Since UniRig emits no labels, each predicted bone inherits the category of its containing GT capsule (Sec.~\ref{sec:exp_setup})---a mapping favorable to UniRig, which is judged on placement alone without risk of category mislabeling.
TailorRig achieves \textbf{92.6\%} recall with only \textbf{7.9\%} off-label bones, versus UniRig's \textbf{42.1\%}/\textbf{25.1\%}, confirming that labels govern which structures are generated.

\begin{figure*}[t]
    \centering
    \includegraphics[width=1\linewidth]{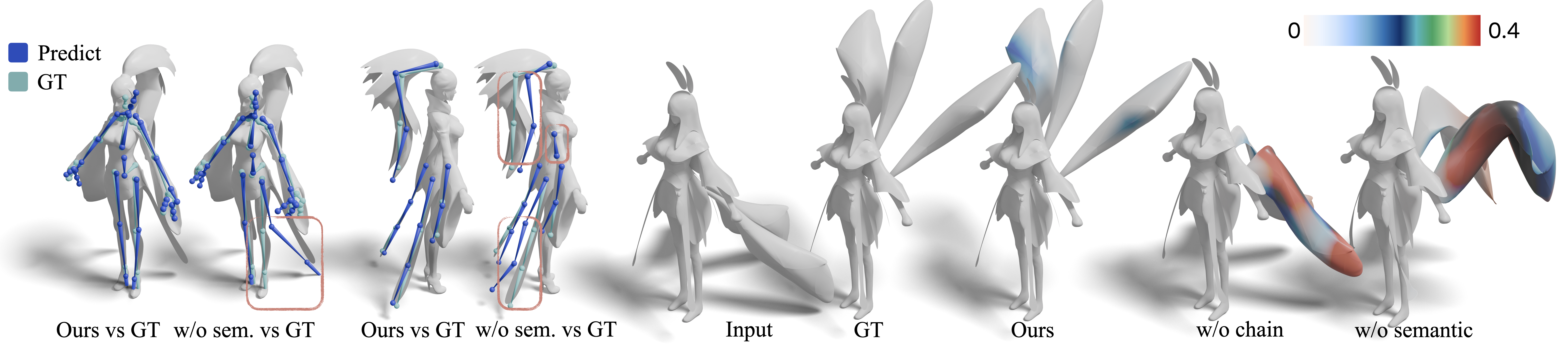}
    \caption{\textbf{Qualitative ablations on skeleton generation and skinning.}
    Left: blue and green denote predicted and ground-truth skeletons, respectively. Removing semantic supervision causes both body and auxiliary bones to drift from their intended regions.
    Right: under articulated deformation, removing the chain-to-bone decomposition or semantic supervision introduces visible cross-part skinning bleeding and deformation artifacts.}
    \Description{Two-panel qualitative ablation. Left: predicted skeletons in blue versus ground-truth skeletons in green, showing bone drift when semantic supervision is removed. Right: deformed meshes showing skinning artifacts when the chain-to-bone decomposition or semantic supervision is removed.}
\label{fig:semantic_ablation}
\end{figure*}

\subsection{Ablation Studies}
\label{sec:struct_val}

We ablate key design choices to validate two forms of structure in our framework: semantic structure (Sec.~\ref{sec:abl_sem}) and geometric-kinematic structure (Sec.~\ref{sec:abl_geo}). All variants share the same training data and schedule; each removes exactly one component. Results are summarized in Table~\ref{tab:ablation_all}.

\subsubsection{Semantic Structure}
\label{sec:abl_sem}

We first ablate the semantic structure used to localize garment-related bones and skinning regions.
Removing $\mathcal{L}_{\text{sem}}$ substantially degrades auxiliary generation (J2J 0.068$\to$0.120, B2B 0.098$\to$0.131) and skinning quality ($L_1$ 7.70$\to$11.50, Def-$L_1$ 0.60$\to$0.83), indicating that vertex-level semantic supervision helps both auxiliary placement and cross-part weight separation.
Removing Label-Aware Attention also hurts auxiliary generation, increasing J2J from 0.068 to 0.107 and B2B from 0.098 to 0.125, showing that category labels need to interact with mesh and body features rather than act as global tags.
Fig.~\ref{fig:semantic_ablation} shows the corresponding failures: skeleton joints drift from the intended regions, and skinning weights bleed across unrelated parts.

\subsubsection{Geometric Structure}
\label{sec:abl_geo}

\paragraph{Hierarchical Bone Encoder.}
\label{sec:abl_hier}
The hierarchical bone encoder architecture is used in all three modules; replacing it with a bone encoder that omits chain information (\textit{w/o Hier.\ Encoder} in Table~\ref{tab:ablation_all}) degrades all three tasks---body J2J rises from 0.023 to 0.026, auxiliary J2J from 0.068 to 0.096, and skinning Def-$L_1$ from 0.60 to 1.88---confirming that chain-level hierarchy provides useful priors throughout the pipeline.
\paragraph{Topology-Aware Body Alignment.}
The \textit{Body Skeleton Alignment} block in Table~\ref{tab:ablation_all} isolates two geometric choices.
Removing TAMB Fusion increases body J2J/B2B from 0.023/0.011 to 0.030/0.015, while removing $\mathcal{L}_{\text{proj}}$ increases body J2J/B2B from 0.023/0.011 to 0.031/0.017, confirming that topology-aware fusion and bone-axis projection provide complementary constraints for body skeleton fitting.
\paragraph{Structural Priors for Skinning Prediction.}
Removing the chain-to-bone decomposition (\textit{w/o Chain-to-Bone}) raises skinning $L_1$ from 7.70 to 11.24 and Def-$L_1$ from 0.60 to 0.84.
Without the coarse chain assignment stage, weights leak across part boundaries, producing visible cross-part bleeding in Fig.~\ref{fig:semantic_ablation} (Right).

\subsection{In-the-Wild Production Evaluation}
\label{sec:in_the_wild}

We further perform evaluation on 50 diverse in-the-wild assets covering different genders, body types, and clothing styles such as layered skirts, ribbons, and capes. Five professional technical artists independently assess each method's output in a blind setting, rating overall rig quality (skeleton and skinning jointly) on a 5-point Likert scale (1: broken, 5: production-ready) and estimating the manual repair time needed to bring the rig to production standard.
The 50 evaluation-only assets come from a held-out internal production pool disjoint from Human-Mix, with 52--128 total bones and 2--36 auxiliary bones per character. Artists specify the auxiliary categories they intend to author; only TailorRig consumes these labels, while all methods are judged against the same intended rig. Method names are hidden and presentation order is randomized per asset. Repair time is an expert estimate rather than a timed editing session. Ratings are highly consistent across artists (ICC(2,k)$=0.95$).

\begin{table*}[t]
\centering
\footnotesize
\setlength{\tabcolsep}{2.8pt}
\renewcommand{\arraystretch}{0.92}
\caption{\textbf{In-the-wild production evaluation} on 50 assets rated by 5 artists.
Qual.: 1--5 Likert; Repair Time: estimated hours to production quality.}
\Description{Table summarizing the in-the-wild production evaluation with Likert quality ratings and estimated repair hours from five artists on fifty assets.}
\label{tab:in_the_wild_user_study}
\begin{tabular}{lccccc}
\toprule
\textbf{Method} & \makecell{\textbf{Body}\\\textbf{J2J}$\downarrow$} & \makecell{\textbf{Aux}\\\textbf{J2J}$\downarrow$} & \makecell{\textbf{Skin}\\\textbf{L1}$\downarrow$} & \textbf{Qual.}$\uparrow$ & \makecell{\textbf{Repair}\\\textbf{Time}$\downarrow$} \\
\midrule
Pinocchio~\cite{baran2007pinocchio} & 0.142 & 0.274 & 0.010 & 2.23 & 3.48 \\
Puppet.~\cite{song2025puppeteer} & 0.082 & 0.226 & 0.033 & 2.98 & 2.70 \\
UniRig~\cite{zhang2025unirig} & 0.025 & 0.188 & 0.010 & 3.37 & 1.38 \\
\textbf{Ours} & \textbf{0.024} & \textbf{0.160} & \textbf{0.003} & \textbf{4.56} & \textbf{0.80} \\
\bottomrule
\end{tabular}
\end{table*}

On in-the-wild assets (Table~\ref{tab:in_the_wild_user_study}), TailorRig attains the lowest errors, the highest artist rating (4.56/5), and the shortest repair time (0.80\,h)---42\% below UniRig (1.38\,h).
On the held-out Human-Mix split, an end-to-end test where each method uses its \emph{own} predicted skeleton and skinning weights, with only bone rotations optimized to fit the same five target poses, shows the same ordering: Def-$L_1$ is $1.16\times10^{-2}$ for TailorRig vs.\ $1.50\times10^{-2}$ for UniRig. This cross-topology test is distinct from the controlled-skeleton comparison in Table~\ref{tab:combined_comp_results}.

\section{Conclusion and Future Work}
We presented TailorRig, a controllable, hierarchical framework for garment-aware rigging that couples template-based body alignment with label-conditioned auxiliary generation and chain-to-bone skinning. It improves template consistency, garment flexibility, and deformation stability for production assets. Current limitations include the fixed discretization used in auxiliary bone generation, which may introduce small offsets in dense regions, and the reliance on a humanoid template with predefined semantic labels.
Future work will explore adaptive or multi-scale representations and extend the taxonomy to broader character types and accessory structures.

\clearpage
\bibliographystyle{ACM-Reference-Format}
\bibliography{refs}

@String{Computing = "Computing" }

@String{Computer = "{IEEE} Computer" }

@article{liu2025riganything,
  title={Riganything: Template-free autoregressive rigging for diverse 3d assets},
  author={Liu, Isabella and Xu, Zhan and Yifan, Wang and Tan, Hao and Xu, Zexiang and Wang, Xiaolong and Su, Hao and Shi, Zifan},
  journal={ACM Transactions on Graphics (TOG)},
  volume={44},
  number={4},
  pages={1--12},
  year={2025},
  publisher={ACM New York, NY, USA}
}

@inproceedings{guo2025mia,
  title={Make-it-animatable: An efficient framework for authoring animation-ready 3d characters},
  author={Guo, Zhiyang and Xiang, Jinxu and Ma, Kai and Zhou, Wengang and Li, Houqiang and Zhang, Ran},
  booktitle={Proceedings of the Computer Vision and Pattern Recognition Conference},
  pages={10783--10792},
  year={2025}
}

@article{xu2020rignet,
  title={Rignet: Neural rigging for articulated characters},
  author={Xu, Zhan and Zhou, Yang and Kalogerakis, Evangelos and Landreth, Chris and Singh, Karan},
  journal={arXiv preprint arXiv:2005.00559},
  year={2020}
}

@article{zhang2025unirig,
  title={One model to rig them all: Diverse skeleton rigging with unirig},
  author={Zhang, Jia-Peng and Pu, Cheng-Feng and Guo, Meng-Hao and Cao, Yan-Pei and Hu, Shi-Min},
  journal={ACM Transactions on Graphics (TOG)},
  volume={44},
  number={4},
  pages={1--18},
  year={2025},
  publisher={ACM New York, NY, USA}
}

@article{song2025puppeteer,
  title={Puppeteer: Rig and animate your 3d models},
  author={Song, Chaoyue and Li, Xiu and Yang, Fan and Xu, Zhongcong and Wei, Jiacheng and Liu, Fayao and Feng, Jiashi and Lin, Guosheng and Zhang, Jianfeng},
  journal={arXiv preprint arXiv:2508.10898},
  year={2025}
}

@inproceedings{song2025magicarticulate,
  title={Magicarticulate: Make your 3d models articulation-ready},
  author={Song, Chaoyue and Zhang, Jianfeng and Li, Xiu and Yang, Fan and Chen, Yiwen and Xu, Zhongcong and Liew, Jun Hao and Guo, Xiaoyang and Liu, Fayao and Feng, Jiashi and others},
  booktitle={Proceedings of the Computer Vision and Pattern Recognition Conference},
  pages={15998--16007},
  year={2025}
}

@article{guo2025autoconnect,
  title={Auto-Connect: Connectivity-Preserving RigFormer with Direct Preference Optimization},
  author={Guo, Jingfeng and Liu, Jian and Chen, Jinnan and Mao, Shiwei and Hu, Changrong and Jiang, Puhua and Yu, Junlin and Xu, Jing and Liu, Qi and Xu, Lixin and others},
  journal={arXiv preprint arXiv:2506.11430},
  year={2025}
}

@article{ma2023tarig,
  title={TARig: Adaptive template-aware neural rigging for humanoid characters},
  author={Ma, Jing and Zhang, Dongliang},
  journal={Computers \& Graphics},
  volume={114},
  pages={158--167},
  year={2023},
  publisher={Elsevier}
}

@inproceedings{chu2025humanrig,
  title={Humanrig: Learning automatic rigging for humanoid character in a large scale dataset},
  author={Chu, Zedong and Xiong, Feng and Liu, Meiduo and Zhang, Jinzhi and Shao, Mingqi and Sun, Zhaoxu and Wang, Di and Xu, Mu},
  booktitle={Proceedings of the Computer Vision and Pattern Recognition Conference},
  pages={304--313},
  year={2025}
}

@article{liu2019neuroskinning,
  title={Neuroskinning: Automatic skin binding for production characters with deep graph networks},
  author={Liu, Lijuan and Zheng, Youyi and Tang, Di and Yuan, Yi and Fan, Changjie and Zhou, Kun},
  journal={ACM Transactions on Graphics (ToG)},
  volume={38},
  number={4},
  pages={1--12},
  year={2019},
  publisher={ACM New York, NY, USA}
}

@article{li2021neuralblendshapes,
  title={Learning skeletal articulations with neural blend shapes},
  author={Li, Peizhuo and Aberman, Kfir and Hanocka, Rana and Liu, Libin and Sorkine-Hornung, Olga and Chen, Baoquan},
  journal={ACM Transactions on Graphics (TOG)},
  volume={40},
  number={4},
  pages={1--15},
  year={2021},
  publisher={ACM New York, NY, USA}
}

@inproceedings{pan2021heterskinnet,
  title={HeterSkinNet: A heterogeneous network for skin weights prediction},
  author={Pan, Xiaoyu and Huang, Jiancong and Mai, Jiaming and Wang, He and Li, Honglin and Su, Tongkui and Wang, Wenjun and Jin, Xiaogang},
  booktitle={Proceedings of the ACM on computer graphics and interactive techniques},
  volume={4},
  number={1},
  year={2021},
  organization={Association for Computing Machinery}
}

@article{baran2007pinocchio,
  title={Automatic rigging and animation of 3d characters},
  author={Baran, Ilya and Popovi{\'c}, Jovan},
  journal={ACM Transactions on graphics (TOG)},
  volume={26},
  number={3},
  pages={72--es},
  year={2007},
  publisher={ACM New York, NY, USA}
}

@inproceedings{dionne2013geodesic,
  title={Geodesic voxel binding for production character meshes},
  author={Dionne, Olivier and de Lasa, Martin},
  booktitle={Proceedings of the 12th ACM SIGGRAPH/Eurographics Symposium on Computer Animation},
  pages={173--180},
  year={2013}
}

@article{zhang2022opt,
  title={Opt: Open pre-trained transformer language models},
  author={Zhang, Susan and Roller, Stephen and Goyal, Naman and Artetxe, Mikel and Chen, Moya and Chen, Shuohui and Dewan, Christopher and Diab, Mona and Li, Xian and Lin, Xi Victoria and others},
  journal={arXiv preprint arXiv:2205.01068},
  year={2022}
}

@inproceedings{liu2025partfield,
  title={Partfield: Learning 3d feature fields for part segmentation and beyond},
  author={Liu, Minghua and Uy, Mikaela Angelina and Xiang, Donglai and Su, Hao and Fidler, Sanja and Sharp, Nicholas and Gao, Jun},
  booktitle={Proceedings of the IEEE/CVF International Conference on Computer Vision},
  pages={9704--9715},
  year={2025}
}

@incollection{lefevre2025comparing,
  title={Comparing 3D Data Representations for Keypoint Estimation in Humanoid Characters},
  author={Lef{\`e}vre, Jeanne-Emma and Cheynel, Th{\'e}o and Daniel, Thomas},
  booktitle={Proceedings of the SIGGRAPH Asia 2025 Posters},
  pages={1--3},
  year={2025}
}

@article{wang2019edgeconv,
  title={Dynamic graph cnn for learning on point clouds},
  author={Wang, Yue and Sun, Yongbin and Liu, Ziwei and Sarma, Sanjay E and Bronstein, Michael M and Solomon, Justin M},
  journal={ACM Transactions on Graphics (tog)},
  volume={38},
  number={5},
  pages={1--12},
  year={2019},
  publisher={Acm New York, NY, USA}
}

@inproceedings{nash2020polygen,
  title={Polygen: An autoregressive generative model of 3d meshes},
  author={Nash, Charlie and Ganin, Yaroslav and Eslami, SM Ali and Battaglia, Peter},
  booktitle={International conference on machine learning},
  pages={7220--7229},
  year={2020},
  organization={PMLR}
}

@inproceedings{siddiqui2024meshgpt,
  title={Meshgpt: Generating triangle meshes with decoder-only transformers},
  author={Siddiqui, Yawar and Alliegro, Antonio and Artemov, Alexey and Tommasi, Tatiana and Sirigatti, Daniele and Rosov, Vladislav and Dai, Angela and Nie{\ss}ner, Matthias},
  booktitle={Proceedings of the IEEE/CVF conference on computer vision and pattern recognition},
  pages={19615--19625},
  year={2024}
}

@inproceedings{yan2022shapeformer,
  title={Shapeformer: Transformer-based shape completion via sparse representation},
  author={Yan, Xingguang and Lin, Liqiang and Mitra, Niloy J and Lischinski, Dani and Cohen-Or, Daniel and Huang, Hui},
  booktitle={Proceedings of the IEEE/CVF conference on computer vision and pattern recognition},
  pages={6239--6249},
  year={2022}
}

@incollection{isozaki2021vroid,
  title={VRoid studio: a tool for making anime-like 3D characters using your imagination},
  author={Isozaki, Nozomi and Ishima, Shigeyoshi and Yamada, Yusuke and Obuchi, Yutaka and Sato, Rika and Shimizu, Norio},
  booktitle={SIGGRAPH Asia 2021 Real-Time Live!},
  pages={1--1},
  year={2021}
}

@inproceedings{mittal2022autosdf,
  title={Autosdf: Shape priors for 3d completion, reconstruction and generation},
  author={Mittal, Paritosh and Cheng, Yen-Chi and Singh, Maneesh and Tulsiani, Shubham},
  booktitle={Proceedings of the IEEE/CVF conference on computer vision and pattern recognition},
  pages={306--315},
  year={2022}
}

@inproceedings{ibing2023octree,
  title={Octree transformer: Autoregressive 3d shape generation on hierarchically structured sequences},
  author={Ibing, Moritz and Kobsik, Gregor and Kobbelt, Leif},
  booktitle={Proceedings of the IEEE/CVF Conference on Computer Vision and Pattern Recognition},
  pages={2698--2707},
  year={2023}
}

@article{hong2023lrm,
  title={Lrm: Large reconstruction model for single image to 3d},
  author={Hong, Yicong and Zhang, Kai and Gu, Jiuxiang and Bi, Sai and Zhou, Yang and Liu, Difan and Liu, Feng and Sunkavalli, Kalyan and Bui, Trung and Tan, Hao},
  journal={arXiv preprint arXiv:2311.04400},
  year={2023}
}

@article{li2023instant3d,
  title={Instant3d: Fast text-to-3d with sparse-view generation and large reconstruction model},
  author={Li, Jiahao and Tan, Hao and Zhang, Kai and Xu, Zexiang and Luan, Fujun and Xu, Yinghao and Hong, Yicong and Sunkavalli, Kalyan and Shakhnarovich, Greg and Bi, Sai},
  journal={arXiv preprint arXiv:2311.06214},
  year={2023}
}

@article{liu2023one,
  title={One-2-3-45: Any single image to 3d mesh in 45 seconds without per-shape optimization},
  author={Liu, Minghua and Xu, Chao and Jin, Haian and Chen, Linghao and Varma T, Mukund and Xu, Zexiang and Su, Hao},
  journal={Advances in Neural Information Processing Systems},
  volume={36},
  pages={22226--22246},
  year={2023}
}

@article{poole2022dreamfusion,
  title={Dreamfusion: Text-to-3d using 2d diffusion},
  author={Poole, Ben and Jain, Ajay and Barron, Jonathan T and Mildenhall, Ben},
  journal={arXiv preprint arXiv:2209.14988},
  year={2022}
}

@article{zhang2024clay,
  title={Clay: A controllable large-scale generative model for creating high-quality 3d assets},
  author={Zhang, Longwen and Wang, Ziyu and Zhang, Qixuan and Qiu, Qiwei and Pang, Anqi and Jiang, Haoran and Yang, Wei and Xu, Lan and Yu, Jingyi},
  journal={ACM Transactions on Graphics (TOG)},
  volume={43},
  number={4},
  pages={1--20},
  year={2024},
  publisher={ACM New York, NY, USA}
}

@inproceedings{xu2022morig,
  title={Morig: Motion-aware rigging of character meshes from point clouds},
  author={Xu, Zhan and Zhou, Yang and Yi, Li and Kalogerakis, Evangelos},
  booktitle={SIGGRAPH Asia 2022 conference papers},
  pages={1--9},
  year={2022}
}

@inproceedings{amenta1998surface,
  title={Surface reconstruction by Voronoi filtering},
  author={Amenta, Nina and Bern, Marshall},
  booktitle={Proceedings of the fourteenth annual symposium on Computational geometry},
  pages={39--48},
  year={1998}
}

@incollection{tagliasacchi2009curve,
  title={Curve skeleton extraction from incomplete point cloud},
  author={Tagliasacchi, Andrea and Zhang, Hao and Cohen-Or, Daniel},
  booktitle={ACM SIGGRAPH 2009 papers},
  pages={1--9},
  year={2009}
}

@book{parent2012computer,
  title={Computer animation: algorithms and techniques},
  author={Parent, Rick},
  year={2012},
  publisher={Newnes}
}

@article{lai2025hunyuan3d,
  title={Hunyuan3D 2.5: Towards High-Fidelity 3D Assets Generation with Ultimate Details},
  author={Lai, Zeqiang and Zhao, Yunfei and Liu, Haolin and Zhao, Zibo and Lin, Qingxiang and Shi, Huiwen and Yang, Xianghui and Yang, Mingxin and Yang, Shuhui and Feng, Yifei and others},
  journal={arXiv preprint arXiv:2506.16504},
  year={2025}
}

@inproceedings{lin2021point2skeleton,
  title={Point2skeleton: Learning skeletal representations from point clouds},
  author={Lin, Cheng and Li, Changjian and Liu, Yuan and Chen, Nenglun and Choi, Yi-King and Wang, Wenping},
  booktitle={Proceedings of the IEEE/CVF conference on computer vision and pattern recognition},
  pages={4277--4286},
  year={2021}
}

@article{james2005skinning,
  title={Skinning mesh animations},
  author={James, Doug L and Twigg, Christopher D},
  journal={ACM Transactions on Graphics (TOG)},
  volume={24},
  number={3},
  pages={399--407},
  year={2005},
  publisher={ACM New York, NY, USA}
}

@article{jacobson2011bounded,
  title={Bounded biharmonic weights for real-time deformation.},
  author={Jacobson, Alec and Baran, Ilya and Popovic, Jovan and Sorkine, Olga},
  journal={ACM Trans. Graph.},
  volume={30},
  number={4},
  pages={78},
  year={2011}
}

@incollection{abdrashitov2023robust,
  title={Robust skin weights transfer via weight inpainting},
  author={Abdrashitov, Rinat and Raichstat, Kim and Monsen, Jared and Hill, David},
  booktitle={SIGGRAPH Asia 2023 Technical Communications},
  pages={1--4},
  year={2023}
}

@inproceedings{huang2025intersection,
  title={Intersection-free Garment Retargeting},
  author={Huang, Zizhou and Ara{\'u}jo, Chrystiano and Kunz, Andrew and Zorin, Denis and Panozzo, Daniele and Zordan, Victor},
  booktitle={Proceedings of the Special Interest Group on Computer Graphics and Interactive Techniques Conference Conference Papers},
  pages={1--12},
  year={2025}
}

@inproceedings{kavan2010fast,
  title={Fast and efficient skinning of animated meshes},
  author={Kavan, Ladislav and Sloan, P-P and O'Sullivan, Carol},
  booktitle={Computer Graphics Forum},
  volume={29},
  number={2},
  pages={327--336},
  year={2010},
  organization={Wiley Online Library}
}

@inproceedings{deng2025anymate,
  title={Anymate: A dataset and baselines for learning 3d object rigging},
  author={Deng, Yufan and Zhang, Yuhao and Geng, Chen and Wu, Shangzhe and Wu, Jiajun},
  booktitle={Proceedings of the Special Interest Group on Computer Graphics and Interactive Techniques Conference Papers},
  pages={1--10},
  year={2025}
}

@misc{adobe2021mixamo,
  author = {{Adobe Inc.}},
  title = {Mixamo},
  year = {2021},
}
\end{document}